\documentclass{article}
\usepackage{ijcai25}

\usepackage{times}
\usepackage{soul}
\usepackage{url}
\usepackage[hidelinks]{hyperref}
\usepackage[utf8]{inputenc}
\usepackage[small]{caption}
\usepackage{graphicx}
\usepackage{amsmath}
\usepackage{amsthm}
\usepackage{booktabs}
\usepackage{algorithm}
\usepackage{algorithmic}
\usepackage[switch]{lineno}

\usepackage{inconsolata}

\usepackage{booktabs} 
\usepackage{color}
\usepackage{xcolor}

\usepackage{tikz}
\usepackage[edges]{forest}
\definecolor{hidden-draw}{RGB}{205, 44, 36}
\definecolor{hidden-blue}{RGB}{194,232,247}
\definecolor{hidden-orange}{RGB}{243,202,120}
\definecolor{hidden-yellow}{RGB}{242,244,193}
\definecolor{tree-level-1}{RGB}{245,20,85}
\definecolor{tree-level-2}{RGB}{246,86,118}
\definecolor{tree-level-3}{RGB}{248,177,193}
\definecolor{tree-leaf}{RGB}{176,230,198}
\usepackage{arydshln}

\newcommand{\citet}[1]{\citeauthor{#1}~\shortcite{#1}}
\newcommand{\citep}[1]{~\cite{#1}}

\title{GUI Agents with Foundation Models: \\A Comprehensive Survey}

\author{
Shuai Wang$^1$\and
Weiwen Liu$^1$\and
Jingxuan Chen$^1$\and
Yuqi Zhou$^2$\and
Weinan Gan$^1$\and \\
Xingshan Zeng$^1$\and
Yuhan Che$^1$\and 
Shuai Yu$^1$\and
Xinlong Hao$^1$\and
Kun Shao$^1$\and \\
Bin Wang$^1$\and
Chuhan Wu$^1$\and 
Yasheng Wang$^1$\and 
Ruiming Tang$^1$\and
Jianye Hao$^1$
\\
\affiliations
$^1$Huawei Noah's Ark Lab $\quad$
$^2$Renmin University of China\\
\emails
\{wangshuai231, liuweiwen8\}@huawei.com
}

\begin{document}

\maketitle

\begin{abstract}
Recent advances in foundation models, particularly Large Language Models (LLMs) and Multimodal Large Language Models (MLLMs), have facilitated the development of intelligent agents capable of performing complex tasks. By leveraging the ability of (M)LLMs to process and interpret Graphical User Interfaces (GUIs), these agents can autonomously execute user instructions, simulating human-like interactions such as clicking and typing. This survey consolidates recent research on (M)LLM-based GUI agents, highlighting key innovations in data resources, frameworks, and applications. We begin by reviewing representative datasets and benchmarks, followed by an overview of a generalized, unified framework that encapsulates the essential components of prior studies, supported by a detailed taxonomy. Additionally, we explore relevant commercial applications. Drawing insights from existing work, we identify key challenges and propose future research directions. We hope this survey will inspire further advancements in the field of (M)LLM-based GUI agents.
\end{abstract}

\section{Introduction}
Graphical User Interfaces (GUIs) are the primary medium through which humans interact with digital devices. From mobile phones to websites, people engage with GUIs daily, and well-designed GUI agents can significantly enhance the user experience. Thus, research on GUI agents has been extensive. However, traditional rule-based and reinforcement learning-based methods struggle with tasks requiring human-like interactions\citep{liu_reinforcement_2018}, limiting their applicability.

\begin{figure}
    \centering
    \includegraphics[width=\linewidth]{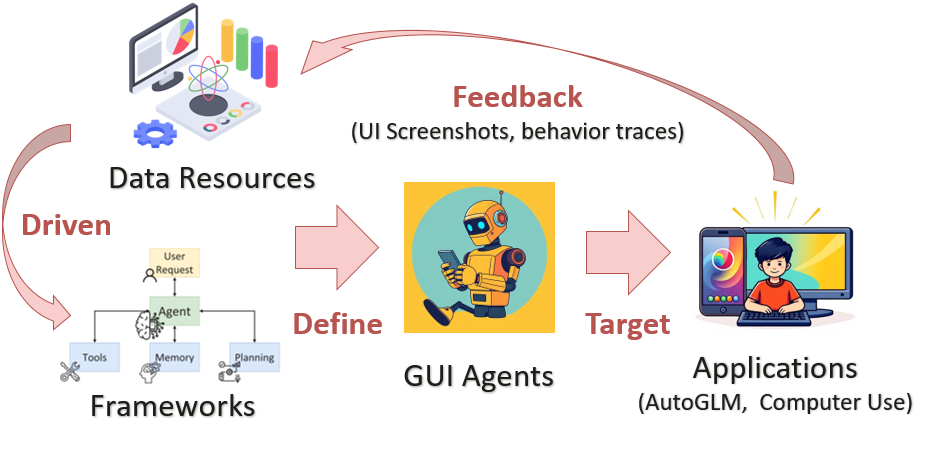}
    \caption{The foundational aspects and goals of GUI agents.}
    \label{fig:idea}
\end{figure}

\begin{figure*}
    \centering
    \includegraphics[width=\linewidth]{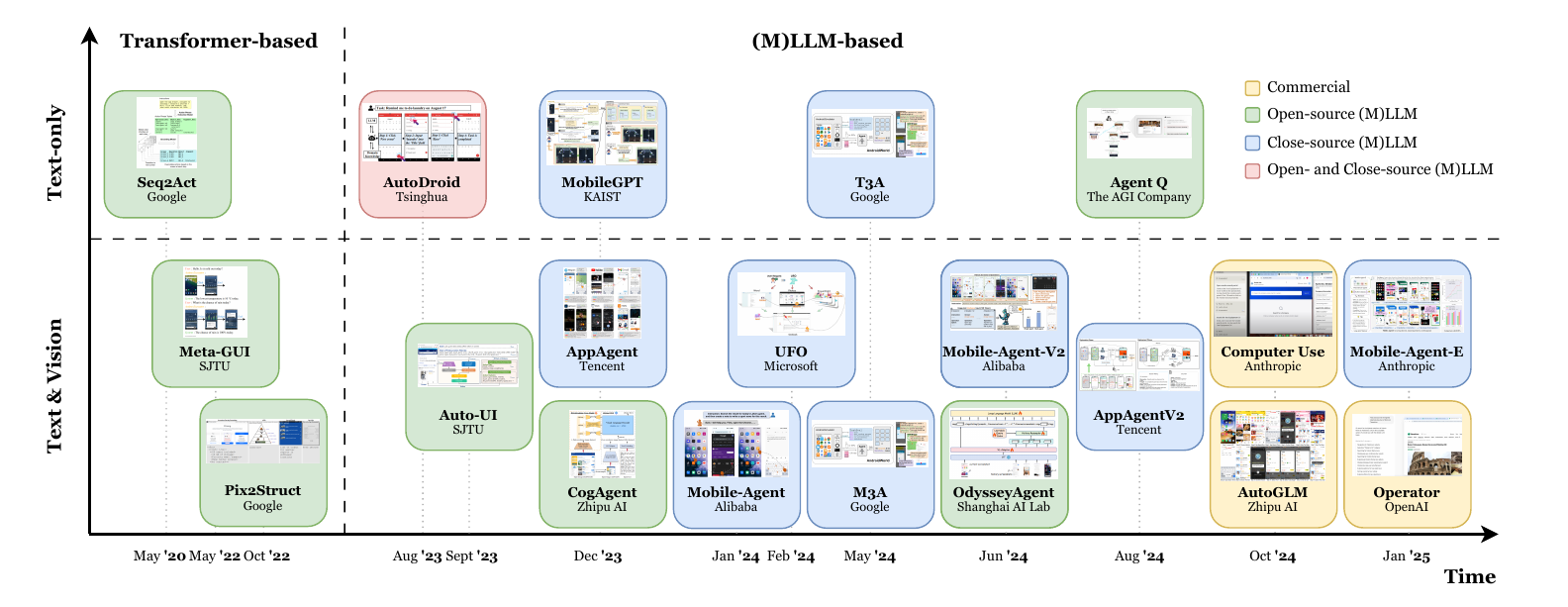}
    \caption{Illustration of the growth trend in the field of GUI agents with foundation models.}
    \label{fig:Summary}
\end{figure*}

Recent advancements in Large Language Models (LLMs) and Multimodal Large Language Models (MLLMs) have significantly enhanced their capabilities in language understanding and cognitive processing~\citep{openai2024gpt4technicalreport,touvron_llama_2023,yang2024qwen2technicalreport}. With improved natural language comprehension and enhanced reasoning abilities, (M)LLM-based agents can now effectively interpret and utilize human language, formulate detailed plans, and execute complex tasks. These breakthroughs provide new opportunities for researchers to address challenges previously considered highly difficult, such as automating tasks within GUIs.

As shown in Figure~\ref{fig:Summary}, recent studies on GUI agents illustrate a shift from simple Transformer-based models to (M)LLM-based agentic frameworks. Their capabilities have expanded from single-modality interactions to multimodal processing, making them increasingly relevant to commercial applications. Given these advancements, we believe it is timely to systematically analyze the development trends of GUI agents, particularly from an application perspective.

This paper aims to provide a structured overview of the latest and influential work in the field of GUI agents. As depicted in Figure~\ref{fig:idea}, we focus on the foundational aspects and goals of GUI agents. Data resources, such as user instructions, User Interface (UI) screenshots, and behavior traces, drive the design of GUI agents\citep{rawles_android_2023,lu2024guiodysseycomprehensivedataset}. Frameworks define the underlying algorithms and models that enable intelligent decision-making\citep{li2024appagent,wang_mobile-agent-v2_2024,zhu2024moba}. Applications represent the optimized and practical goals\citep{lai2024autowebglm,liu2024autoglm}. The current state of these aspects reflects the maturity of the field and highlights future research priorities.

To this end, we organize this survey around three key areas: \textbf{Data Resources}, \textbf{Frameworks}, and \textbf{Applications}. The main contributions of this paper are: 1) a comprehensive summary of existing research and a detailed review of current data sources, providing a useful guide for newcomers to the field; 2) a unified and generalized GUI agent framework with clearly defined and categorized functional components to facilitate a structured review; 3) an analysis of trends in both research and commercial applications of GUI agents. 

\section{GUI Agent Data Resources}
Recent research has focused on developing datasets and benchmarks to train and evaluate the capabilities of (M)LLM-based GUI agents. A variety of datasets are available for training GUI agents. These agents employ different approaches to interact with environments. Additionally, multiple methods have been proposed for evaluation.

\textbf{Dataset:} 
Common datasets for training GUI agents typically contain natural language instructions that describe task goals, along with demonstration trajectories that include screenshots and action pairs. A pioneering work in this area is PIXELHELP~\citep{li2020mappingnaturallanguageinstructions}, which introduces a new class of problems focused on translating natural language instructions into actions on mobile user interfaces. In recent years, Android in the Wild~\citep{rawles_android_2023} has created a dataset featuring a variety of single-step and multi-step tasks. Aimed at advancing GUI navigation agent research, Android-In-The-Zoo~\citep{zhang2024androidzoochainofactionthoughtgui} introduces a benchmark dataset with chained action reasoning annotations.

Insight-UI~\citep{shen2024falcon} automatically constructs a GUI pre-training dataset that simulates multiple platforms across 312,000 domains. To assess model performance both within and beyond the scope of training data, AndroidControl~\citep{li2024effects} includes demonstrations of daily tasks along with both high- and low-level human-generated instructions. The scope of mobile control datasets is further extended from single-application to cross-application scenarios by GUI-Odyssey~\citep{lu2024guiodysseycomprehensivedataset}.

Most of the aforementioned datasets are primarily limited to English and image-based tasks. However, UGIF Dataset~\citep{venkatesh2024ugif} covers eight languages, Mobile3M~\citep{wu2024mobilevlm} focuses on Chinese, and GUI-WORLD~\citep{chen2024guiworlddatasetguiorientedmultimodal} includes video annotations, expanding the dataset landscape for broader multilingual and multimodal research.

\textbf{Environment:} 
GUI agents require environments for task execution, which can be broadly categorized into three types. The first category is static environments, where the environment remains fixed as it was when developed. Agents in this category operate within predefined datasets without the ability to create new states.

In contrast, the second and third categories involve dynamic environments, where new outcomes can emerge during agent execution. The key distinction between these categories lies in whether the dynamic environment is simulated or realistic. Simulations of real-world environments require additional implementation but are often cleaner and free of distractions, such as pop-ups and advertisements. WebArena~\citep{zhou_webarena_2023} implements a versatile website covering e-commerce, social forums, collaborative software development, and content management. Similarly, GUI Testing Arena~\citep{zhao2024gui} provides a standardized environment for testing GUI agents, including defect injection.

For realistic environments, agents interact directly with web or mobile platforms as human users do, better reflecting real-world conditions. SPA-Bench~\citep{chen2024spa} encompasses tasks that involve both system and third-party mobile applications, supporting single-app and cross-app scenarios in both English and Chinese.

\tikzstyle{my-box}=[
    rectangle,
    draw=hidden-draw,
    rounded corners,
    text opacity=1,
    minimum height=1.5em,
    minimum width=5em,
    inner sep=2pt,
    align=center,
    fill opacity=.5,
]
\tikzstyle{leaf}=[my-box, minimum height=1.5em,
    fill=hidden-orange!60, text=black, align=left,font=\scriptsize,
    inner xsep=2pt,
    inner ysep=4pt,
]

\begin{figure*}[ht]
    \centering
    \resizebox{\textwidth}{!}{
        \begin{forest}
            forked edges,
            for tree={
                grow=east,
                reversed=true,
                anchor=base west,
                parent anchor=east,
                child anchor=west,
                base=left,
                font=\small,
                rectangle,
                draw=hidden-draw,
                rounded corners,
                align=left,
                minimum width=4em,
                edge+={darkgray, line width=1pt},
                s sep=3pt,
                inner xsep=2pt,
                inner ysep=3pt,
                ver/.style={rotate=90, child anchor=north, parent anchor=south, anchor=center},
            },
            where level=1{text width=4.3em,font=\scriptsize,}{},
            where level=2{text width=5.8em,font=\scriptsize,}{},
            where level=3{text width=6.1em,font=\scriptsize,}{},
            where level=4{text width=6.1em,font=\scriptsize,}{},
            [  (M)LLM-based \\ GUI Agent 
                [
                    (M)LLM-based \\ GUI Agent \\ Framework 
    \\ (Section \ref{sec:construction})
                    [
                        GUI Perceiver \\ 
                        [
                            Wen~\citep{wen_empowering_2023,wen_droidbot-gpt_2024}{,} Li~\citep{li2020mappingnaturallanguageinstructions}{,} Ferret-{UI}~\citep{you_ferret-ui_2024}{,} \\ Omniparser~\citep{lu2024omniparser}{,} Iris~\citep{ge2024iris} Ferret-{UI}~\citep{you_ferret-ui_2024}{,} \\
                                , leaf, text width=33.6em
                        ]
                    ]
                    [
                        Task Planner \\ 
                        [
                                UFO~\citep{zhang_ufo_2024}{,} Mobile-Agent-V2~\citep{wang_mobile-agent-v2_2024}{,} AppAgent~\citep{zhang_appagent_2023}
                                , leaf, text width=33.6em
                        ]
                    ]
                    [
                        Decision Maker \\ 
                        [
                                GUI Odyssey~\citep{lu2024guiodysseycomprehensivedataset}{,} UFO~\citep{zhang_ufo_2024}{,} GUI AutoDroid~\citep{wen2024autodroidllmpoweredtaskautomation}{,} \\ MobileAgent~\citep{ding_mobileagent_2024}
                                , leaf, text width=33.6em
                        ]
                    ]
                    [
                        Executor \\ 
                        [
                                AndroidWorld~\citep{rawles_androidworld_2024}
                                , leaf, text width=33.6em
                        ]
                    ]
                    [
                        Memory Retriever \\ 
                        [
                                Wang~\citep{wang_survey_2024}{,} GUI Odyssey~\citep{lu2024guiodysseycomprehensivedataset}{,} AppAgent~\citep{zhang_appagent_2023}{,} \\ MobileAgent~\citep{ding_mobileagent_2024}{,} Agent Q~\citep{putta2024agent}
                                , leaf, text width=33.6em
                        ]
                    ]
                ]
                [
                    (M)LLM-based \\ GUI Agent \\ Taxonomy \\ (Section \ref{sec:taxonomy})
                    [
                        Input Modality \\
                        [
                            LLM-based \\
                            [
                                Lee~\citep{lee_explore_2023}{,} 
                                Li~\citep{li2020mappingnaturallanguageinstructions}{,}
                                WebGPT~\citep{nakano_webgpt_2022}{,} \\
                                Laser~\citep{ma2023laser}{,}
                                AutoWebGLM~\citep{lai2024autowebglm}{,}
                                Agent Q~\citep{putta2024agent}
                                , leaf, text width=26em
                            ]
                        ]
                        [
                            MLLM-based \\
                            [
                                UIBert~\citep{bai_uibert_2021}{,}
                                Zhang~\citep{zhang_screen_2021}{,}
                                Pix2Struct~\citep{lee_pix2struct_2023}{,} \\
                                Aria-UI~\citep{yang2024aria}{,}
                                Ferret-UI~\citep{you_ferret-ui_2024}{,}
                                Moba~\citep{zhu2024moba}{,} \\
                                AutoGLM~\citep{liu2024autoglm}{,}
                                Tars-UI~\citep{qin2025ui}{,}
                                AppAgent~\citep{zhang_appagent_2023}{,} \\
                                AndroidWorld~\citep{rawles_androidworld_2024}{,}
                                Mobile-Agent-V2~\citep{wang_mobile-agent-v2_2024}{,} \\
                                Mobile-Agent-E~\citep{ding_mobileagent_2024}{,}
                                SeeAct~\citep{zheng2024gpt4visiongeneralistwebagent}{,}
                                 \\
                                Spotlight~\citep{li_spotlight_2023}{,}
                                MobileVLM~\citep{wu2024mobilevlm}{,}\\
                                OdysseyAgent~\citep{lu2024guiodysseycomprehensivedataset} 
                                AppAgent V2~\citep{li2024appagent}{,}
                                , leaf, text width=26em
                            ]
                        ]
                    ]
                    [
                        Learning Mode \\
                        [
                            Prompting-based \\
                            [
                                AppAgent~\citep{zhang_appagent_2023}{,}
                                AppAgent V2~\citep{li2024appagent}{,} \\
                                Mobile-Agent-V2~\citep{wang_mobile-agent-v2_2024}{,}
                                Mobile-Agent-E~\citep{ding_mobileagent_2024}{,} \\
                                Wen~\citep{wen_droidbot-gpt_2024}{,} 
                                OpenAgents~\citep{xie2024openagents}{,}
                                UFO~\citep{zhang_ufo_2024}{,} \\
                                WebVoyager~\citep{he2024webvoyager}{,}
                                Moba~\citep{zhu2024moba}{,}
                                Yan~\citep{yan_gpt-4v_2023}{,} \\
                                Zheng~\citep{zheng2024gpt4visiongeneralistwebagent}{,}
                                Wang~\citep{wang_enabling_2023}
                                , leaf, text width=26em
                            ]
                        ]
                        [
                            Tuning-based \\
                            [  Kil~\citep{kil2024dual}{,}
                                MobileVLM~\citep{wu2024mobilevlm}{,}
                               Furuta~\citep{furuta_multimodal_2023}{,}
                                 \\
                               Wang~\citep{lee_explore_2023}{,}
                               MobileAgent~\citep{ding_mobileagent_2024}{,}
                               {META}-{GUI}~\citep{sun_meta-gui_2022}{,}\\
                               Kim~\citep{kim_language_2023}{,}
                               PC Agent~\citep{he2024pc}{,}
                               OdysseyAgent~\citep{lu2024guiodysseycomprehensivedataset}{,}
                                \\
                               Zhang~\citep{zhang_you_2023}{,}
                               Aria-UI~\citep{yang2024aria}{,} \\
                               Aguvis~\citep{xu2024aguvis}{,} 
                               WebGPT~\citep{nakano_webgpt_2022}{,}
                               SeeAct~\citep{zheng2024gpt4visiongeneralistwebagent}{,}\\
                               Wen~\citep{wen_empowering_2023}{,}
                               AutoGLM~\citep{liu2024autoglm}{,}
                               ScreenAgent~\citep{niu2024screenagent}
                                , leaf, text width=26em
                            ]
                        ]
                    ]
                ]
            ]
        \end{forest}
    }
    \caption{A comprehensive taxonomy of (M)LLM-based GUI Agents: frameworks, modality, and learning paradigms.}
    \label{overall}
\end{figure*}

\textbf{Evaluation:} 
Another critical component of GUI agent datasets is the evaluation of agent performance. The most common and important metric is success rate, which measures how effectively an agent completes tasks. Additional metrics, such as efficiency, are sometimes considered as well.

Evaluation methods are often closely tied to the environment type. In static environments, action matching is a widely used method that compares an agent’s executed action sequence with a human demonstration (e.g.,~\citet{rawles_android_2023},~\citet{li2024effects}). However, a major limitation of action matching is its inability to account for multiple successful execution paths, leading to false negatives when evaluating agent performance.

Evaluating dynamic environments, whether simulated or realistic, presents additional challenges due to their uncertain conditions. Evaluation methods can range from fully human-dependent to semi-automated and fully automated approaches. Human evaluations require manual verification, making them non-reusable. In AppAgent~\citep{li2024appagent} and MobileAgent~\citep{ding_mobileagent_2024}, human evaluators assess whether each agent-executed task was successful. Semi-automated evaluations involve human-developed validation logic that can be reused for different execution trajectories of the same task. For example, WebArena~\citep{zhou_webarena_2023} and AndroidWorld~\citep{rawles_androidworld_2024} incorporate handcrafted validation functions for task completion. Fully automated evaluations eliminate human involvement by relying on models for success detection. SPA-Bench~\citep{chen2024spa}, for instance, employs MLLMs for evaluating task completion. Although reducing human labor is crucial for large-scale evaluation, balancing efficiency with accuracy remains a key research challenge.

\section{(M)LLM-based GUI Agent}
With the human-like capabilities of (M)LLMs, GUI agents aim to handle various tasks to meet users' needs. Organizing the frameworks of GUI agents and designing methods to optimize their performance is crucial to unlocking the full potential of (M)LLMs. As shown in Figure~\ref{overall}, we summarize a generalized \textbf{Framework} and discuss its components in relation to existing works in Section~\ref{sec:construction}. Building on this foundation, we then review recent influential \textbf{Methods} for constructing and optimizing GUI agents, categorizing them with an exhaustive taxonomy in Section~\ref{sec:taxonomy}.

\begin{figure*}
    \centering
    \includegraphics[width=0.9\linewidth]{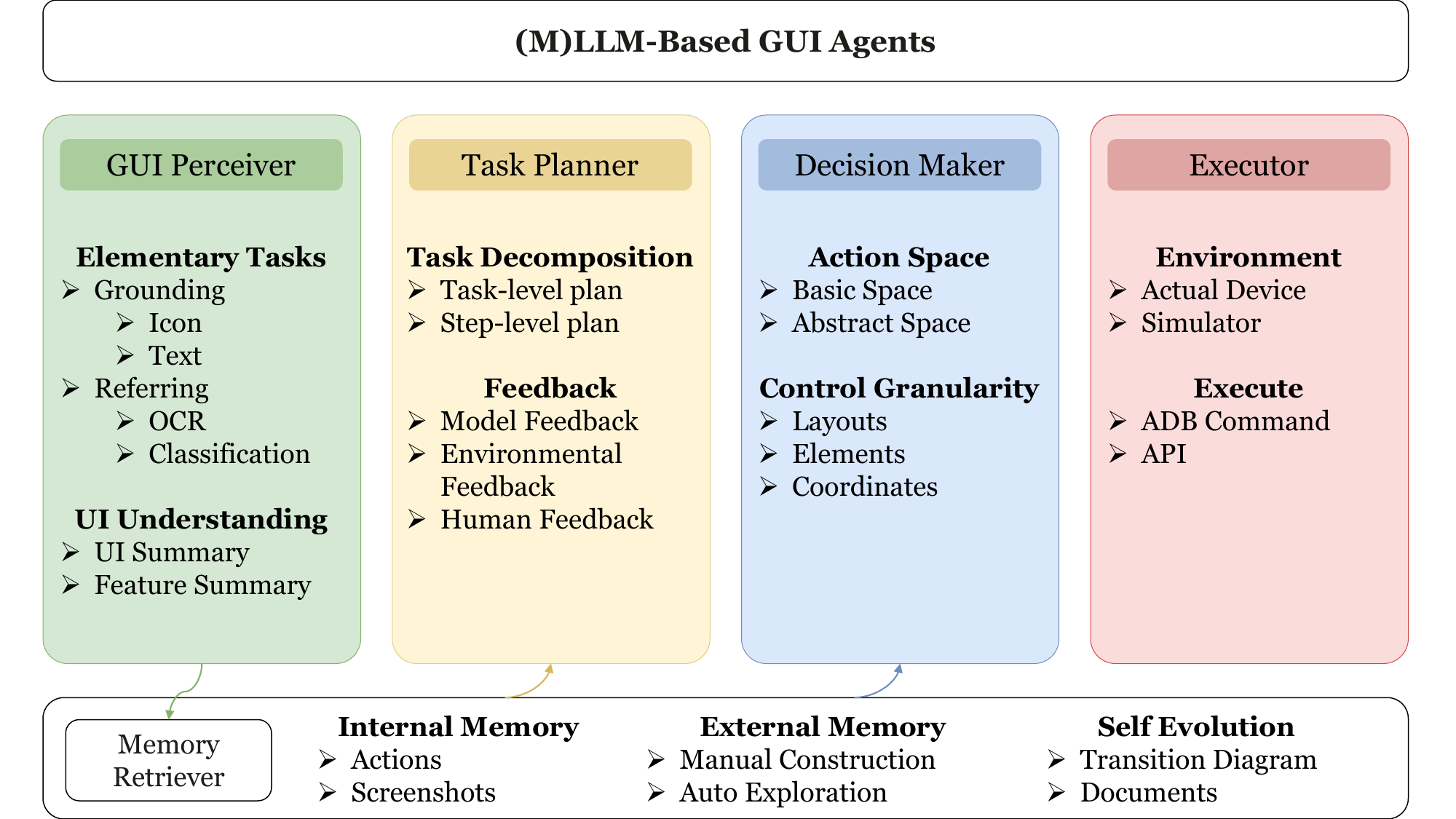}
    \caption{(M)LLM-based GUI agents: the generalized framework and key technologies.}
    \label{fig:overall_framework}
    \vspace{-7pt}
\end{figure*}

\subsection{(M)LLM-based GUI Agent Framework}~\label{sec:construction}

The goal of GUI agents is to automatically control a device to complete tasks defined by the user. Typically, GUI agents take a user's query and the device's UI status as inputs and generate a series of human-like actions to achieve the tasks.

As shown in Figure~\ref{fig:overall_framework}, we present a generalized (M)LLM-based GUI agent framework, consisting of five components: GUI Perceiver, Task Planner, Decision Maker, Memory Retriever, and Executor. Many variations of this framework exist. For instance, ~\citet{wang_mobile-agent-v2_2024} proposes a multi-agent GUI control framework comprising a planning agent, a decision agent, and a reflection agent to tackle navigation challenges in mobile device operations. This approach shares functional similarities with our proposed framework. A follow-up study by~\citet{wang2025mobile} further disentangles high-level planning from low-level action decisions by employing dedicated agents and introduces memory-based self-evolution to enhance performance.

\textbf{GUI Perceiver:} To effectively complete a device task, a GUI agent should accurately interpret user input and detect changes in the device's UI. Although language models excel in understanding user intent~\citep{touvron_llama_2023,openai2024gpt4technicalreport}, navigating device UIs requires a reliable visual perception model to understand GUIs.

A GUI Perceiver appears explicitly or implicitly in GUI agent frameworks. For agents based on single-modal LLMs~\citep{wen_empowering_2023,wen_droidbot-gpt_2024,li2020mappingnaturallanguageinstructions}, a GUI Perceiver is usually an explicit module of the frameworks. However, for agents with multi-modal LLMs~\citep{hong_cogagent_2023,zhang_appagent_2023,wang2024mobileagentautonomousmultimodalmobile}, UI perception is seen as a capability of the model itself.

UI perception is also an important problem in GUI agent research, some work\citep{you_ferret-ui_2024,zhang_screen_2021,lu2024omniparser} focuses on understanding and processing UIs, rather than building the agent. For example, Pix2struct\citep{lee_pix2struct_2023} employs a ViT-based image-encoder-text-decoder architecture, which pre-trains on Screenshot-HTML data pairs and fine-tunes for specific tasks. This method has shown strong performance in web-based visual comprehension tasks. Screen2words\citep{wang_screen2words_2021} is a novel approach that encapsulates a UI screen into a coherent language representation, which is based on a transformer encoder-decoder architecture to process UIs and generate the representation.  To address the defects of purely vision-based screen parsing methods, \citet{ge2024iris} introduces Iris, a visual agent for GUI understanding, addressing challenges related to architectural limitations for heterogeneous GUI information and annotation bias in GUI training via two innovations: An information-sensitive architecture to prioritize high-density UI regions via edge detection, and a dual-learning strategy that refines visual/functional knowledge iteratively using unlabeled data, reducing annotation dependence. 

\textbf{Task Planner:} The GUI agent should effectively decompose complex tasks, often employing a Chain-of-Thought (CoT) approach. Due to the complexity of tasks, recent studies~\citep{zhang_ufo_2024,wang_mobile-agent-v2_2024} introduce an additional module to support more detailed planning.

Throughout the GUI agent's process, plans may adapt dynamically based on decision feedback, typically achieved through a ReAct-style. For instance, \citet{zhang_appagent_2023} uses on-screen observations to enhance the CoT for improved decision-making, while \citet{wang_mobile-agent-v2_2024} develops a reflection agent that provides feedback to refine plans.

\textbf{Decision Maker:} A Decision Maker provides the next operation(s) to control a device. Most studies\citep{lu2024guiodysseycomprehensivedataset,zhang_ufo_2024,wen2024autodroidllmpoweredtaskautomation} define a set of UI-related actions—such as click, text, and scroll—as a basic action space. In a more complicated case, \citet{ding_mobileagent_2024} encapsulates a sequence of actions to create Standard Operating Procedures(SOPs) to guide further operations.

As the power of GUI agents improves, the granularity of operations becomes more refined. Recent work has progressed from element-level operations\citep{zhang_appagent_2023,wang2024mobileagentautonomousmultimodalmobile} to coordinate-level controls\citep{wang_mobile-agent-v2_2024,hong_cogagent_2023}.

\textbf{Executor:} An Executor maps outputs to the relevant environments. While most studies use Android Debug Bridge (ADB) to control real devices \citep{li2024appagent,wang_mobile-agent-v2_2024}, ~\citet{rawles_androidworld_2024} develops a simulator to access additional UI-related information.

\textbf{Memory Retriever:}  A Memory Retriever is designed as an additional source of information to help agents perform tasks more effectively~\citep{wang_survey_2024}. 

GUI agents' memory is typically divided into internal and external categories. Internal memory~\citep{lu2024guiodysseycomprehensivedataset} consists of prior actions, screenshots, and system states during execution, while external memory~\citep{zhang_appagent_2023,ding_mobileagent_2024} includes knowledge and rules related to the UI or task, providing additional inputs for the agent. 

\subsection{(M)LLM-based GUI Agent Taxonomy}\label{sec:taxonomy}
Consequently, this paper classifies existing work with the difference of input modality and learning mode in Figure~\ref{overall}. 

\subsubsection{GUI Agents with Different Input modality}
\textbf{LLM-based GUI Agents:}
With the limited multimodal capability, earlier GUI agents~\citep{lee_explore_2023,li2020mappingnaturallanguageinstructions,ma2023laser,lai2024autowebglm,putta2024agent,nakano_webgpt_2022,nakano_webgpt_2022} often require a GUI perceiver to convert GUI screens into text-based inputs. 

So, parsing and grounding the GUI screens is the first step. For instance,~\citet{li2020mappingnaturallanguageinstructions} transforms the screen into a series of object descriptions and applies a transformer-based action mapping. The problem definitions and datasets have spurred further research.  ~\citet{you_ferret-ui_2024} proposes a series of referring and grounding tasks, which provide valuable insights into the pre-training of GUIs.~\citet{lu2024omniparser} proposes a screen parsing framework incorporating the local semantics of functionality with interactable region detection for better UI understanding and element grounding. 

Afterward, LLMs are used as the core of the agents. ~\citet{wen2024autodroidllmpoweredtaskautomation} further converts GUI screenshots into a simplified HTML representation for compatibility with the LLMs. By combining GUI representation with app-specific knowledge, they build Auto-Droid, a GUI agent based on online GPT and on-device Vicuna.  In the field of web automation, LASER~\citep{ma2023laser} navigates web environments purely through text, treating web navigation as state-space exploration to enable flexible state transitions and error recovery. Similarly, AutoWebGLM~\citep{lai2024autowebglm} processes HTML text data without visual inputs, refining webpage structures to preserve key information for ChatGLM3-6B. Agent Q~\citep{putta2024agent} further extends this paradigm by relying solely on HTML DOM text for reasoning and decision-making, emphasizing language models for planning and action execution. WebGPT~\citep{nakano_webgpt_2022}, a fine-tuned GPT-3 model, uses text-based web browsing (processing HTML content) to collect information via commands like searching and clicking. It generates answers supported by references and is optimized using human feedback and rejection sampling.

\noindent
\textbf{MLLM-based GUI Agents:}
Recent studies~\citep{wang_mobile-agent-v2_2024,bai_uibert_2021,zhang_appagent_2023,kim_language_2023} utilize the multimodal capabilities of advanced (M)LLMs to improve GUI comprehension and task execution.

Leveraging the visual understanding capabilities of MLLMs, recent studies~\citep{wang_mobile-agent-v2_2024,li_spotlight_2023,bai_uibert_2021,zhu2024moba,qin2025ui} explore end-to-end frameworks for GUI device control. For example, Spotlight~\citep{li_spotlight_2023} proposes a Vision-Language model framework, pre-trained on Web/mobile data and fine-tuned for UI tasks. This model greatly improves the ability to understand UIs. By combining screenshots with a user focus as input, Spotlight outperforms previous methods on multiple UI understanding tasks, showing verified gains in downstream tasks. Likewise, VUT~\citep{li_vut_2021} is proposed for GUI understanding and multi-modal UI input modeling, using two Transformers: one for encoding and fusing image, structural, and language inputs, and the other for linking three task heads to complete five distinct UI modeling tasks and learn downstream multiple tasks end-to-end. Experiments show that VUT's multi-task learning framework can achieve state-of-the-art (SOTA) performance on UI modeling tasks. UIbert~\citep{bai_uibert_2021} focuses on heterogeneous GUI features and considers that the multi-modal information in the GUI is self-aligned. UIbert is a transformer-based joint image-text model, which is pre-trained in large-scale unlabeled GUI data to learn the feature representation of UI elements.~\citet{zhu2024moba} presents a two-level agent structure for executing complex and dynamic GUI tasks. Moba's Global Agent handles high-level planning, while the Local Agent selects actions for sub-tasks, streamlining the decision-making process with improved efficiency. UI-TARS~\citep{qin2025ui} navigates interfaces through screenshots, enabling human-like interactions via keyboard and mouse. Leveraging a large-scale GUI dataset, it achieves context-aware UI understanding and precise captioning.

To enhance performance, some studies~\citep{zhang_appagent_2023,rawles_androidworld_2024} utilize additional invisible metadata. For instance, AndroidWorld~\citep{rawles_androidworld_2024} establishes a fully functional Android environment with real-world tasks, serving as a benchmark for evaluating GUI agents. They propose M3A, a zero-shot prompting agent that uses Set-of-Marks as input. Experiments with M3A variants assess how different input modalities—text, screenshots, and accessibility trees—affect GUI agent performance.~\citet{yang2024aria} proposes a framework incorporating dynamic action history with both textual and interleaved text-image formats, which allows it to ground elements more effectively for dynamic, multi-step scenarios. 

\subsubsection{GUI Agents with Different Learning Mode}

\begin{table*}[htbp]
    \centering
    \caption{Overview of (M)LLM-Based GUI Agents.}
    ~\label{tab:model_comparison}
    \resizebox{\textwidth}{!}{
    \begin{tabular}{@{}lcccccc@{}}
        \toprule
        \textbf{Model Name} & \textbf{Category} & \textbf{GUI Perceiver} & \textbf{Learning Method} & \textbf{Base Model} & \textbf{Scenarios} \\
        \midrule
        \multicolumn{6}{c}{\textbf{Prompting-based}} \\
        \midrule
        PaLM~\citep{wang_enabling_2023} & Single Step & HTML & Few-shot prompting & PaLM & Mobile \\
        MM-Navigator~\citep{yan_gpt-4v_2023} & Single Step & Screenshot & Zero-shot prompting & GPT-4V & Mobile \\
        \hdashline
        MemoDroid~\citep{lee_explore_2023} & End-to-End & HTML & Few-shot prompting & ChatGPT/GPT-4V & Mobile/Desktop \\
        AutoTask~\citep{pan_autotask_2023} & End-to-End & Screenshot/API & Zero-shot prompting & GPT-4V & Mobile \\
        AppAgent~\citep{zhang_appagent_2023} & End-to-End & Screenshot & Exploration-based/In-context learning & GPT4V & Mobile \\
        DroidBot-GPT~\citep{wen_droidbot-gpt_2024} & End-to-End & Screenshot & Zero-shot prompting & ChatGPT & Mobile \\
        Mobile-Agent-V2~\citep{wang_mobile-agent-v2_2024} & End-to-End & Screenshot & Zero-shot prompting & GPT4V & Mobile \\
        SeeAct~\citep{zheng2024gpt4visiongeneralistwebagent} & End-to-End & Screenshot/HTML &  Few-shot prompting & GPT-4V & Web\\
        Mobile-Agent-E~\citep{wang2025mobile} & End-to-End & Screenshot & Zero-shot prompting & GPT-4o/Claude-3.5-Sonnet/Gemini-1.5-pro & Mobile \\
        \midrule
        \multicolumn{6}{c}{\textbf{Learning-based}} \\
        \midrule
        Spotlight~\citep{li_spotlight_2023} & UI modeling & Screenshot & Pretrain/SFT & ViT & Mobile/Web \\
        Pix2Struct~\citep{lee_pix2struct_2023} & UI modeling & Screenshot & Pretrain/SFT & ViT & Web \\
        VUT~\citep{li_vut_2021} & UI modeling & Screenshot & SFT & Transformer & Mobile/Web \\
        Screen Recognition~\citep{zhang_screen_2021} & UI modeling & Screenshot & SFT & Faster R-CNN & Mobile \\
        Screen2Words~\citep{wang_screen2words_2021} & UI modeling & Screenshot & SFT & Transformer & Mobile \\
        Aria-UI~\citep{yang2024aria} & UI modeling & Screenshot & Pretrain/SFT & Aria & Mobile/Web/Desktop \\
        Ferret-UI~\citep{you_ferret-ui_2024} & UI modeling & Screenshot & Pretrain/SFT  & Ferret & Mobile \\
        \hdashline
        AutoDroid~\citep{wen2024autodroidllmpoweredtaskautomation} & End-to-End & HTML & Exploration-based/SFT & Vicuna-7B & Mobile \\
        Seq2Act~\citep{li2020mappingnaturallanguageinstructions} & End-to-End & Texts & Supervised learning & Transformer & Mobile \\
        Meta-GUI~\citep{sun_meta-gui_2022} & End-to-End & Screenshot/XML & Supervised learning & Transformer & Mobile \\
        Agent Q~\citep{putta2024agent} & End-to-End & Screenshot/DOM & RL/BC Training & Transformer & Web \\
        WebGUM~\citep{furuta_multimodal_2023} & End-to-End & Screenshot/HTML & SFT & Flan-T5 & Web \\
        CogAgent~\citep{hong_cogagent_2023} & End-to-End & Screenshot & SFT & CogVLM & Mobile/Desktop \\
        MobileVLM~\citep{wu2024mobilevlm} & End-to-End & XML/Screenshot & Pretrain/SFT & Qwen-VL-Chat & Mobile \\
        WebGPT~\citep{nakano_webgpt_2022} & End-to-End & Texts & SFT & GPT-3 & Web \\
        AutoGLM~\citep{liu2024autoglm} & End-to-End & Screenshot/HTML & Pretrain/SFT/RL & ChatGLM & Mobile/Web \\
        OdysseyAgent~\citep{lu2024guiodysseycomprehensivedataset} & End-to-End & Screenshot & SFT & Qwen-VL & Mobile \\
        \bottomrule
    \end{tabular}
    }
\end{table*}

\textbf{Prompting-based GUI Agents:}
Prompting is an effective approach to building agents with minimal extra computational overhead. Given the diversity of GUIs and tasks, numerous studies~\citep{zhang_appagent_2023,li2024appagent,wang_mobile-agent-v2_2024,wen_droidbot-gpt_2024,xie2024openagents,zhang_ufo_2024,he2024webvoyager} use prompting to create GUI agents, adopting CoT or ReAct styles.

Recent studies use prompting to build and simulate the functions of GUI agent components. For example,~\citet{yan_gpt-4v_2023} introduces MM-Navigator, which utilizes GPT-4V for zero-shot GUI understanding and navigation. For the first time, this work demonstrates the significant potential of LLMs, particularly GPT-4V, for zero-shot GUI tasks. Manual evaluations show that MM-Navigator achieves impressive performance in generating reasonable action descriptions and single-step instructions for iOS tasks. Additionally,~\citet{wen_droidbot-gpt_2024} presents DroidBot-GPT, which summarizes the app's status, past actions, and tasks into a prompt, using ChatGPT to choose the next action. Beyond mobile applications, prompting-based approaches have also been widely adopted in web-based GUI agents. ~\citet{zheng2024gpt4visiongeneralistwebagent} proposes SeeAct, a GPT-4V-based generalist web agent. With screenshots as input, SeeAct generates action descriptions and converts them into executable actions with designed action grounding techniques. OpenAgents~\citep{xie2024openagents} leverages prompts to guide browser extensions in executing tasks such as web navigation and form filling, operating purely on the reasoning capabilities of LLMs without additional training. Similarly, WebVoyager~\citep{he2024webvoyager} integrates visual and textual information from screenshots and web pages, using prompts to interpret UI elements and execute interactions like clicking and typing. UFO~\citep{zhang_ufo_2024} dynamically generates task plans and executes actions through prompting, allowing it to generalize across diverse web tasks without requiring task-specific adaptations.

Some studies enable the GUI agent to fully leverage external knowledge through prompting to complete GUI tasks.

AppAgent~\citep{zhang_appagent_2023} proposes a multi-modal agent framework to simulate human-like mobile phone operations. The framework is divided into two phases: Exploration, where agents explore applications and document their operations, and Deployment, where these documents guide the agent in observing, thinking, acting, and summarizing tasks. This is the first work to claim human-like GUI automation capabilities. AppAgent V2~\citep{li2024appagent} further improves GUI parsing, document generation, and prompt integration by incorporating optical character recognition (OCR) and detection tools, moving beyond the limitations of off-the-shelf parsers for UI element identification.~\citet{wang_enabling_2023} uses a pure in-context learning method to implement interaction between LLMs and mobile UIs. The method divides the conversations between agents and users into four categories from the originator and designs a series of structural CoT prompting to adapt an LLM to execute mobile UI tasks. MobileGPT~\citep{lee_explore_2023} emulates the cognitive processes of human use of applications to enhance the LLM-based agent with a human-like app memory. MobileGPT uses a random explorer to explore and generate screen-related subtasks on many apps and save them as app memory. During the execution, the related memory is recalled to complete tasks. 

\noindent
\textbf{SFT-based GUI Agents:}
Supervised fine-tuning (SFT) allows (M)LLMs to adapt to specific domains and perform customized tasks with high efficiency. Recent studies on GUI agents~\citep{wen_empowering_2023,furuta_multimodal_2023,niu2024screenagent,he2024pc,kil2024dual} demonstrate the benefits of SFT for GUI agents to process new modal inputs, learn specific procedures, or execute specialized tasks.

For instance,~\citet{furuta_multimodal_2023} proposes WebGUM for web navigation. WebGUM is jointly fine-tuned with an instruction-optimized language model and a vision encoder, incorporating temporal and local perceptual capabilities. The evaluation results on MiniWoB show that WebGUM outperforms GPT-4-based agents. ~\citet{zhang_you_2023} introduces Auto-UI, a multimodal solution combining an image-language encoder-decoder architecture with a Chain of Actions policy, fine-tuned on the AitW dataset. This Chain of Actions captures intermediate previous action histories and future action plans.~\citet{yang2024aria} proposes a data-centric pipeline to generate high-quality generalization data from publicly available data. This data is used to fine-tune the VLM for diverse instructions in various environments.~\citet{xu2024aguvis} introduces a two-stage training paradigm for AGUVIS. In the first stage, the agent learns visual representations of GUI components through self-supervised learning. In the second stage, it fine-tunes interactive tasks using reinforcement learning, enabling efficient autonomous GUI interaction. On computer-based environments, ScreenAgent~\citep{niu2024screenagent} fine-tunes the ScreenAgent dataset, mapping screenshots to action sequences. It operates via VNC, following a planning-acting-reflecting framework inspired by Kolb's experiential learning. PC-Agent~\citep{he2024pc} employs a multi-agent architecture, fine-tuning a planning agent on cognitive trajectories collected via PC Tracker, enabling it to model human cognitive patterns. Additionally,~\citet{kil2024dual} fine-tunes DeBERTa for element ranking and Flan-T5 for action prediction, incorporating visual signals to enhance web navigation.

In summary, we provide a systematic overview of recent influential research on (M)LLM-based GUI agents. We address their goal formulations, input perceptions, and learning paradigms, as shown in Table~\ref{tab:model_comparison}

\section{Industrial Applications of (M)LLM-Based GUI Agents}

GUI agents have been widely used in industrial settings, such as mobile assistants and search agents, demonstrating significant commercial value and potential.

\paragraph{Google Assistant for Android:}  By saying phrases like ``Hey Google, start a run on Example App,'' users can use Google Assistant for Android to launch apps, perform tasks, and access content. App Actions, powered by built-in intents (BIIs), enhance app functionality by integrating with Google Assistant. This enables users to navigate apps and access features through voice queries, which the Assistant interprets to display the desired screen or widget.

\paragraph{Apple Intelligence:}
Apple Intelligence is the suite of AI-powered features and services developed by Apple. This includes technologies such as machine learning, natural language processing, and computer vision that power features like Siri, facial recognition, and photo organization. Apple also integrates AI into its hardware and software ecosystem to improve device performance and user experience. Their focus on privacy means that much of this AI processing happens on-device, ensuring that user data remains secure.

\paragraph{New Bing:}
Microsoft's search engine is designed to offer users a more intuitive, efficient, and comprehensive search experience. Leveraging cutting-edge artificial intelligence and machine learning technologies, New Bing goes beyond traditional keyword searches to understand the context and intent behind user queries. With New Bing as an example, the LLM-based deep search engine is also an important form of GUI agents.

\paragraph{Anthropic Computer Use:} Anthropic's ``Computer Use'' feature enables Claude to interact with tools and manipulate a desktop environment. By understanding and executing commands, Computer-Using Agent(CUA) can perform the necessary actions to complete tasks, much like a human.

\paragraph{OpenAI Operator: } OpenAI recently introduced Operator, an AI agent capable of autonomously performing tasks using its own browser. This agent leverages the CUA model, which combines GPT-4o's vision capabilities with advanced reasoning through reinforcement learning. Operator can interpret screenshots and interact with GUIs—such as buttons, menus, and text fields—just as humans do. This development marks a significant advancement in AI capabilities, enabling more efficient and autonomous interactions with digital interfaces.

\paragraph{Microsoft Copilot:}
An AI tool in Microsoft 365 apps for productivity with GPT-based suggestions, task automation, and content generation. Enhances workflows, creativity, and decision-making with real-time insights.

\paragraph{AutoGLM:}
AutoGLM\citep{liu2024autoglm} is designed for autonomous mission completion via GUIs on platforms like phones and the web. Its Android capability allows it to understand user instructions autonomously without manual input, enabling it to handle complex tasks such as ordering takeout, editing comments, shopping, and summarizing articles.

\paragraph{MagicOS 9.0 YOYO:}
An advanced assistant with four main features: natural language and vision processing, user behavior learning, intent recognition and decision-making, and seamless app integration. It understands user habits to autonomously fulfill requests, such as ordering coffee through voice commands, by navigating apps and services.
\section{Challenges}
Due to the rapid development of this field, we summarize several key research questions that require urgent attention: 

\paragraph{Personalized GUI Agents:} Due to the personal nature of user devices, GUI agents inherently interact with personalized information. As an example, users may commute from home to work during weekdays, while walking to their favorite restaurants and cafes on weekends. The integration of personalized information would clearly enhance the user experience with GUI agents. As the capabilities of (M)LLMs continue to improve, personalized GUI agents have become a priority. Effectively collecting and utilizing personal information to deliver a more intelligent experience for users is an essential topic for future research and applications.

\paragraph{Security of GUI Agents:} GUI devices play a crucial role in modern life, making the idea of allowing GUI agents to take control a significant concern for users. For instance, improper operations in financial apps could lead to substantial financial losses, while inappropriate comments on social media apps could damage one's reputation and privacy. Ensuring that GUI agents are not only highly efficient and capable of generalizing but also uphold user-specific security and provide transparency about their actions is an urgent research challenge.  This is a critical issue, as it directly impacts the viability of applying GUI agents in real-world scenarios.

\paragraph{Inference Efficiency:} Humans are highly sensitive to GUI response time, which significantly impacts the user experience. Current (M)LLM-based GUI agents still face notable drawbacks with inference latency. Additionally, communication delay is also an important consideration in real-world applications. As a result, efficient device-cloud collaboration strategies and effective device-side (M)LLM research will become critical areas of focus in the future.

\section{Conclusion}
In this paper, we provide a comprehensive review of the rapidly evolving field of (M)LLM-based GUI Agents. The review is organized into three main perspectives: Data Resources, Frameworks, and Applications. Additionally, we present a detailed taxonomy that connects existing research and highlights key techniques. We also discuss several challenges and propose potential future directions for GUI Agents that leverage foundation models.

\newpage
\bibliographystyle{named}
\bibliography{ijcai25}

\end{document}